\documentclass[10pt,twocolumn,letterpaper]{article}

\usepackage[pagenumbers]{wacv} 

\usepackage{graphicx}
\usepackage{amsmath}
\usepackage{amssymb}
\usepackage{booktabs}
\usepackage{threeparttable}
\usepackage{enumitem}

\makeatletter
\renewcommand\@fnsymbol[1]{}
\makeatother

\usepackage[pagebackref,breaklinks,colorlinks]{hyperref}

\usepackage[capitalize]{cleveref}
\crefname{section}{Sec.}{Secs.}
\Crefname{section}{Section}{Sections}
\Crefname{table}{Table}{Tables}
\crefname{table}{Tab.}{Tabs.}

\def\wacvPaperID{} 
\def\confName{WACV}
\def\confYear{2025}

\newcommand{\Gx}{G_\mathcal{X}}
\newcommand{\Dy}{D_\mathcal{Y}}
\newcommand{\Gy}{G_\mathcal{Y}}
\newcommand{\Dx}{D_\mathcal{X}}

\begin{document}

\title{Recoverable Anonymization for Pose Estimation: A Privacy-Enhancing Approach}


\author{Wenjun Huang$^{1}$ \quad Yang Ni$^{1}$ \quad Arghavan Rezvani$^{1}$ \quad SungHeon Jeong$^{1}$ \\ \quad Hanning Chen$^{1}$ \quad Yezi Liu$^{1}$ \quad Fei Wen$^{2}$ \quad Mohsen Imani$^{1}$ 
\vspace{0.3em} 
\\
{\normalsize $^1$ University of California, Irvine, USA} \quad
{\normalsize $^2$ Texas A\&M University, USA} \\
{\tt\small \{wenjunh3, m.imani\}@uci.edu}
\thanks{This work was supported in part by the DARPA Young Faculty Award, the National Science Foundation (NSF) under Grants \#2127780, \#2319198, \#2321840, \#2312517, and \#2235472, the Semiconductor Research Corporation (SRC), the Office of Naval Research through the Young Investigator Program Award, and Grants \#N00014-21-1-2225 and N00014-22-1-2067. Additionally, support was provided by the Air Force Office of Scientific Research under Award \#FA9550-22-1-0253, along with generous gifts from Xilinx and Cisco.}
}

\maketitle

\begin{abstract}
Human pose estimation (HPE) is crucial for various applications
. However, deploying HPE algorithms in surveillance contexts raises significant privacy concerns due to the potential leakage of sensitive personal information (SPI) such as facial features, and ethnicity. Existing privacy-enhancing methods often compromise either privacy or performance, or they require costly additional modalities.
We propose a novel privacy-enhancing system that generates privacy-enhanced portraits while maintaining high HPE performance. 
Our key innovations include the reversible recovery of SPI for authorized personnel and the preservation of contextual information.
By jointly optimizing a privacy-enhancing module, a privacy recovery module, and a pose estimator, our system ensures robust privacy protection, efficient SPI recovery, and high-performance HPE.
Experimental results demonstrate the system's robust performance in privacy enhancement, SPI recovery, and HPE.
\end{abstract} 
\vspace{-3em}
\section{Introduction}\label{sec:introduction}
With the progression of computer vision, human pose estimation (HPE) has become a crucial and fundamental issue, attracting considerable scholarly attention.
As a pivotal element of human-centric visual understanding, HPE establishes the groundwork for numerous advanced computer vision tasks, such as human action recognition \cite{wang2013action}, human parsing \cite{ruan2019devil}, motion prediction and retargeting \cite{liu2021motion, kappel2021high}. 
Consequently, it underpins a broad collection of applications, including human behavior analysis \cite{teepe2022towards}, violence detection \cite{garcia2023human}, crowd riot scene identification \cite{yogameena2017computer}, and autonomous driving \cite{xu2021action}.

\begin{figure}[t]
  \centering
   \includegraphics[width=\columnwidth]{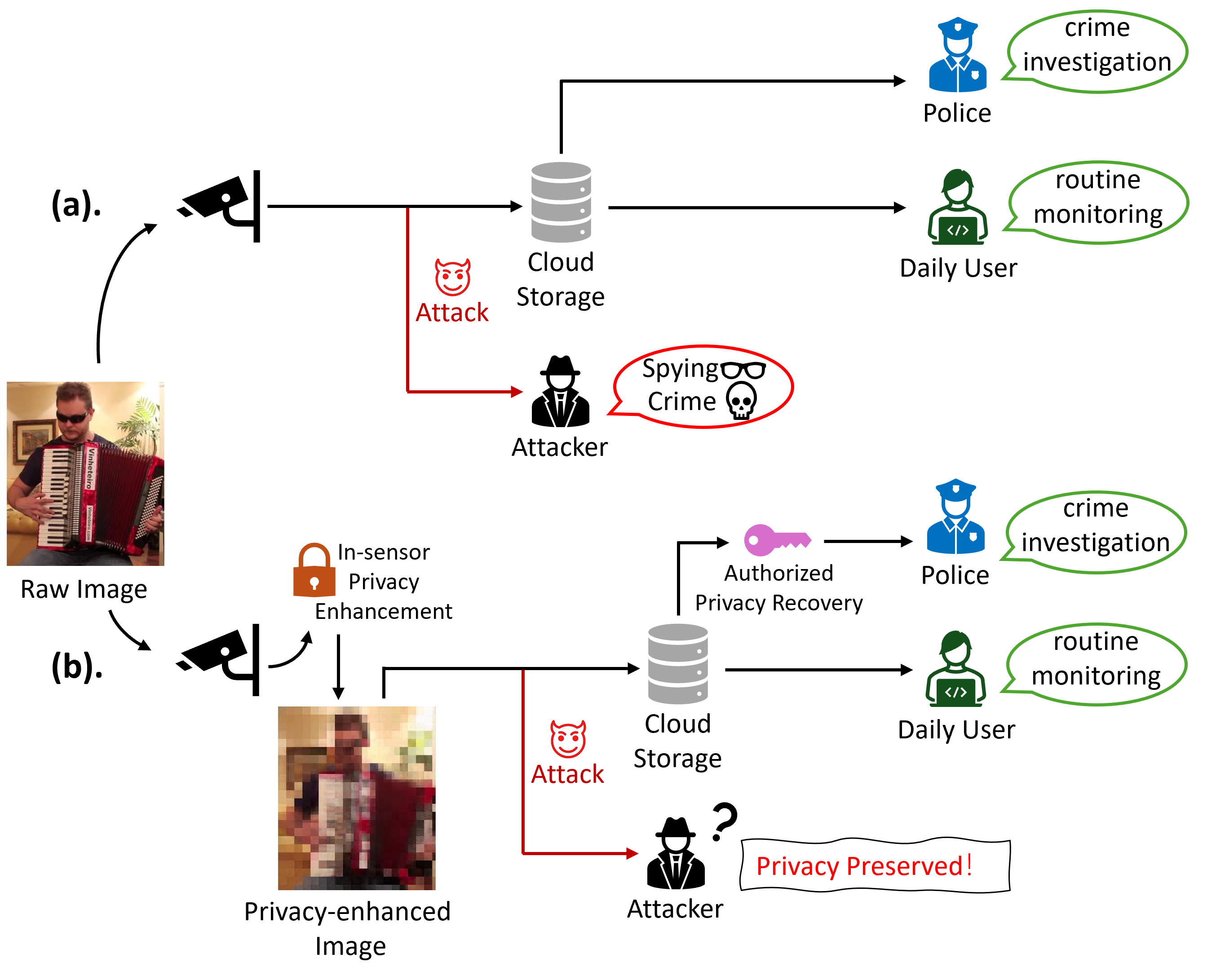}
   \caption{Motivation for our privacy-enhancing system. \textbf{(a).} Conventional surveillance systems are susceptible to leaks of SPI, which can be exploited for illicit surveillance and criminal activities. \textbf{(b).} Our system not only safeguards SPI against information misuse but also supports HPE. The privacy-enhanced images retain functionality for routine monitoring, while SPI remains recoverable by authorized personnel.
   }
   \label{fig:overview}
      \vspace{-2em}
\end{figure}

Due to the extensive computation involved in the applications above, users typically resort to cloud services for data processing and machine learning \cite{yu2024credit, yu2024advanced, zheng2024advanced, huang2021exploration}. 
However, when data is transmitted to cloud servers, sensitive personal information (SPI) such as facial features, gender, and ethnicity is inevitably shared. 
Privacy issues are particularly pronounced in surveillance contexts where HPE algorithms are widely deployed, as illustrated in \cref{fig:overview}\textbf{(a)}. 
Ubiquitous surveillance systems collect and share vast amounts of data.
While this data is valuable for legitimate users in various scenarios, such as routine monitoring, and crime investigations, it simultaneously raises significant privacy concerns for individuals and public safety. 
Without careful protection measures, SPI in raw data could be leaked and misused by malicious parties for harmful purposes. 
For instance, attackers might recognize individuals and surveil them for further criminal activities or even forge their identities \cite{chadha2021deepfake}. 
Additionally, the leakage of SPI can introduce bias and compromise the fairness of analyses and judicial processes \cite{dressel2021dangers}.


In response to data misuse, various legal regulations have been introduced~\cite{barnoviciu2019gdpr,chik2013singapore}, and researchers are developing more advanced algorithms to consider personal privacy. 
For privacy enhancement in computer vision applications, a straightforward solution is to use very low-resolution data \cite{ryoo2017privacy, liu2020indoor}. 
Although these methods do not require specialized training to remove privacy features, they often fail to balance privacy enhancement and model performance effectively.
Some approaches \cite{srivastav2019human, cao2022bed, ahmad2024event} employ additional modalities to enhance privacy.
However, the need to install sensors for these extra modalities increases the cost of surveillance systems, impeding their widespread deployment. 
Another set of methods involves modifying images with hand-crafted features such as blurring, adding noise, and pixelation \cite{agrawal2011person, chen2007tools, padilla2015visual}. 
Unfortunately, these techniques demand extensive domain knowledge, which may not be practical in real-world applications.


Recent privacy-enhancing systems adopt data-driven approaches that conceal SPI from various perspectives. 
For instance, Hukkel{\aa}s \etal \cite{hukkelaas2023deepprivacy2} propose a framework using a generative adversarial network (GAN) for full-body synthesis. 
Their approach generates new representations of individuals that effectively obscure SPI while preserving essential pose information.
In another approach, Hinojosa \etal \cite{hinojosa2021learning} introduces a hardware/software co-design framework. 
This framework optimizes both the point spread function of the camera lens and the neural network architecture, enabling the development of domain-specific computational cameras tailored for privacy-enhancing purposes.
Furthermore, Dave \etal \cite{dave2022spact} present a training framework that autonomously removes SPI in a self-supervised manner, alleviating the need for extensive manual labeling efforts.
Kansal \etal \cite{kansal2024privacy} propose a novel dual-stage framework that suppresses SPI from the discriminative features, and introduces a controllable privacy mechanism through differential privacy.

However, most of the previous work does not target HPE.
Besides, all the aforementioned methods exhibit shortcomings in one or more of the following aspects:

\noindent\textbf{(1). Recovery of Removed SPI:} Privacy-enhanced images should allow authorized users to recover SPI when necessary. 
While SPI may not be essential for scientific research or routine monitoring, it remains critical for specific applications. 
To ensure data utility for various users, authorized personnel such as law enforcement officials should be able to recover original raw images from privacy-enhanced versions, particularly for investigative purposes.

\noindent\textbf{(2). Preservation of Context:} Effective privacy-enhancing systems should modify only the region of interest (e.g., humans) while preserving the background unchanged. 
Contextual information is crucial as the interpretation of actions can vary significantly depending on the surroundings \cite{fiske2020social, gifford2007environmental, chen2024taskclip, yun2024missiongnn}. 
For instance, distinguishing between someone jogging in a park and someone fleeing a store after theft requires intact contextual clues.
Therefore, the context information should be preserved after privacy enhancement, to aid correct interpretation.

\noindent\textbf{(3). Lightweight Deployment:} Privacy-enhancing systems need to be lightweight for deployment near cameras. 
Transmitting data to cloud servers poses security risks such as interception and tampering during transmission \cite{ebadinezhad2023systematic, sarkar2022secure}. 
Deploying privacy-enhancing systems near cameras reduces these vulnerabilities by processing raw images locally before transmission \cite{yun2024hypersense, huang2024intelligent, huang2024ecosense}. 
Therefore, such systems must operate efficiently in real-time, considering limited computational resources and power constraints.


\begin{figure*}[t]
  \centering
   \includegraphics[width=0.8\linewidth]{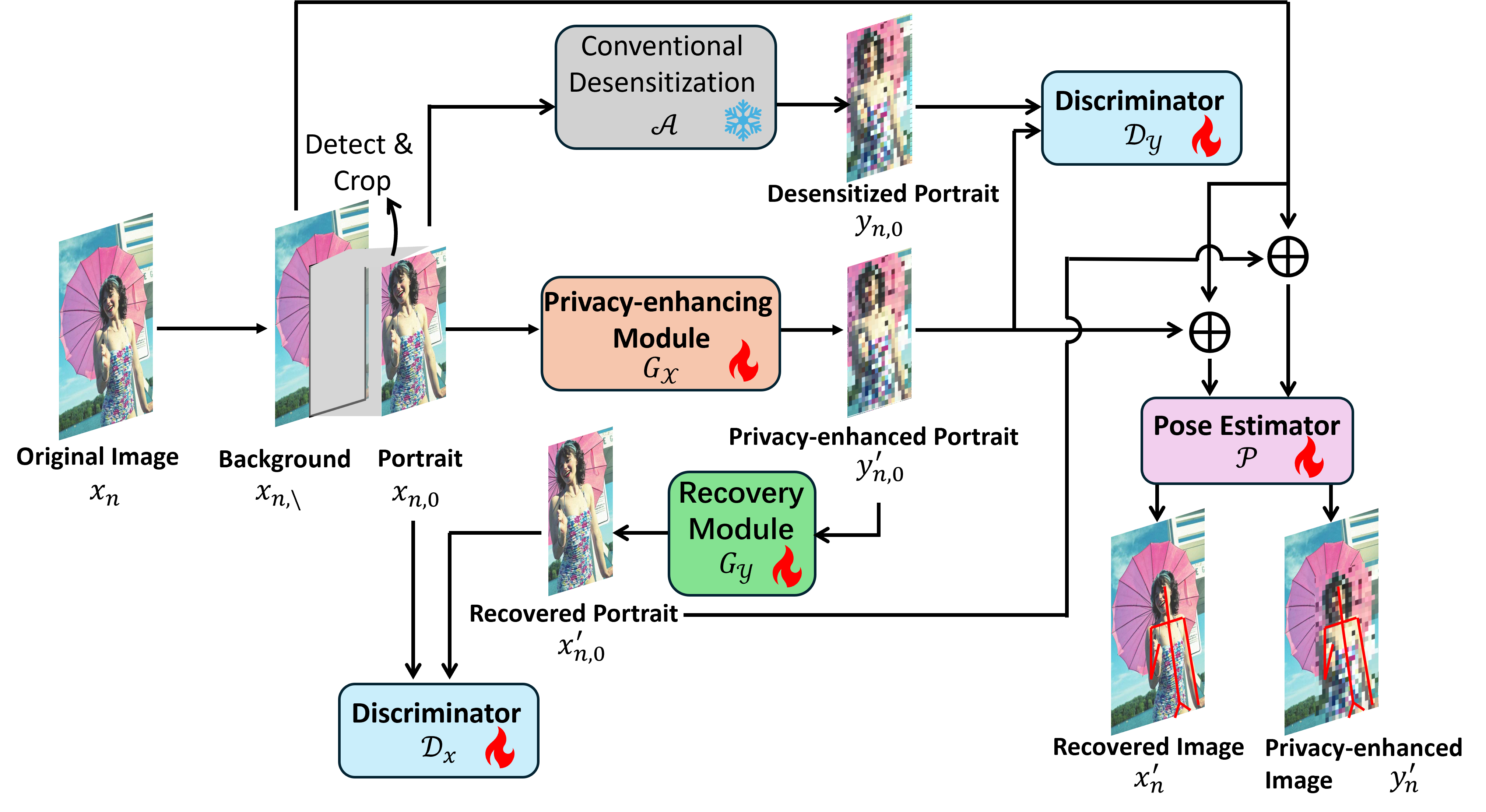}

   \caption{
   The complete pipeline of our proposed system. It contains a privacy-enhancing module $\Gx$ erasing private information, a module $\Gy$ recovering the removed private information, two discriminators $\Dx, \Dy$ for distinguishing the generated portraits, and a pose estimator $\mathcal{P}$ implementing pose estimation. \includegraphics[width=0.023\textwidth]{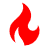} denotes the trainable modules, and \includegraphics[width=0.025\textwidth]{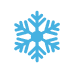}
   denotes the frozen modules.
   }
   \label{fig:framework}
      \vspace{-1em}
\end{figure*}

By addressing the limitations observed in previous work, we propose a privacy-enhancing system capable of generating privacy-enhanced portraits of individuals in images with minimal impact on HPE, as depicted in \cref{fig:overview}\textbf{(b)}. 
The privacy-enhancing module operates near the camera, processing raw images before transmission. 
This approach ensures that SPI in the privacy-enhanced images remains concealed from potential attackers, yet remains usable for HPE tasks and recoverable by authorized users through a privacy recovery module.
Our approach begins by desensitizing raw images using conventional methods such as blurring, pixelation, or noise addition.
These desensitized images serve as initial supervised inputs for the privacy-enhancing module, which then modifies original images to create privacy-enhanced versions in a trainable manner.
To ensure the preservation of essential features for recovery and HPE, we optimize the privacy-enhancing process in conjunction with a privacy recovery model and a pose estimator. 
Through supervised and joint learning, our system achieves effective privacy protection, robust recovery capabilities, and maintains high performance in HPE tasks.
The key contributions of this work are outlined as follows:

\begin{itemize}[leftmargin=0.35cm, itemindent=.0cm, itemsep=0.0cm, topsep=0.1cm]
    \item To the best of our knowledge, we are the first to discuss reversibility, privacy recovery, and context preservation in privacy enhancement for HPE. We introduce a novel privacy-enhancing system designed to generate privacy-enhanced portraits of individuals in images, specifically adapted for downstream machine learning tasks such as HPE.
    \item We proposed an end-to-end joint learning policy for obfuscation, recovery, and pose estimation modules, with the ultimate aim of maintaining pose information and HPE performance after obfuscation and recovery.
    \item We experimentally show that our system achieves robust performance in privacy protection, recovery of original images, and accurate human pose estimation. With joint training, on privacy-enhanced images, our model achieves around 10\% higher average precision than the one that only finetunes the HPE model, while also equipped with strong obfuscation capability. On recovered images, our model further enhances the quality by around 3\%, thanks to the accurate recovery and adaptive injection of HPE-related information.
\end{itemize}


\section{Method}\label{sec:method}

In this section, we elaborate on each component of our proposed system.
As illustrated in \cref{fig:framework}, our system is composed of three modules: 
\textbf{(1). A privacy-enhancing module} (\cref{sec:privacy_enhancing_module}).
We leverage an image-to-image style translation model using conditional generative adversarial networks (cGANs) \cite{mirza2014conditional} to generate privacy-enhanced portraits. 
The privacy-enhanced module is able to anonymize SPI in the images while preserving the features for the downstream tasks. 
The style translation is learned with the guide of a pose estimator such that necessary features are injected for downstream tasks.
\textbf{(2). A privacy recovery module }(\cref{sec:privacy_recovery_module}).
In order to facilitate the reversibility given authorization, we use another pair of cGANs and jointly optimize them with the privacy-enhancing module to recover the SPI.
\textbf{(3). A pose estimator }for human detection and pose estimation on both privacy-enhanced and recovered images (\cref{sec:pose_estimator}).
All modules are tuned end-to-end to maintain pose estimation quality, where the pose estimator provides feedback for the first two modules. 
\subsection{Privacy Enhancing Module}\label{sec:privacy_enhancing_module}
\vspace{-0.7em}

Consider a set of images in the original domain $\{X_0, X_1, \cdots, X_n\}\in\mathcal{X}$. Each image $X_n$ contains one or multiple people of portraits $\{x_{n,0}, x_{n,1}, \cdots, x_{n,i}\} \in \mathcal{X}$ with articulated pose annotations $\{p^x_{n,0}, p^x_{n,1}, \cdots, p^x_{n,i}\}$, where $i$ denotes the portrait index in $X_n$.
We leverage a pretrained lightweight object detector to detect all people and crops the regions to construct a data pool of $\{x_{0,0}, x_{0,1}, \cdots, x_{n,i}\} \in \mathcal{X}$ with poses $\{p_{0,0}^x, p_{0,1}^x, \cdots p_{n,i}^x\}$.

The goal of the privacy-enhancing module can be defined as follows:
Given the pool of training articulated portraits $\{x_{0,0}, \cdots, x_{n,i}\}$ with poses $\{p_{0,0}^x, \cdots p_{n,i}^x\}$, we want to generate the paired privacy-enhanced portraits $\{y_{0,0}^\prime, \cdots, y_{n,i}^\prime\} \in \mathcal{Y}$ in the desensitized domain with poses $\{p_{0,0}^{y^\prime}, \cdots p_{n,i}^{y^\prime}\}$. $y_{n,i}^\prime$ should maintain a high pose feature similarity with the paired portrait $x_{n,i}$ (i.e., $p_{n,i}^x\approx p_{n,i}^{y^\prime}$) while removing the SPI in it.
To achieve this in a learnable manner, we introduce a generator $\Gx$ and discriminator $\Dy$.

The generator $\Gx$ generates the privacy-enhanced portrait $y_{n,i}^\prime = \Gx(x_{n,i})$.
To facilitate the generation, a discriminator $\Dy$ is adopted to learn to distinguish the generated portraits $y_{n,i}^\prime$ and the desensitization style guidance portraits $y_{n,i}=\mathcal{A}(x_{n,i})$, where $y_{n,i}$ is the privacy-enhanced portrait generated from a conventional desensitization method $\mathcal{A}$.
Mathematically, $\Dy$ distinguishes the portrait pair $(y_{n,i}, y_{n,i}^\prime)$ via a discriminator loss:
\begin{align}
    \mathcal{L}_{\Dy} = & -\mathbb{E}_{(x, y) \sim p_{\text{data}}(x, y)}[\log \Dy(y | x)] \notag \\
    & - \mathbb{E}_{x \sim p_{\text{data}}(x)}[\log(1 - \Dy(\Gx(x) | x))] \label{eq:Dy_loss}
\end{align}
, where $\Dy(a | b)$ is the discriminator's output probability that the $a$ is real given the condition $b$.

On the other hand, $\Gx$ tries to trick $\Dy$.
Therefore, it is optimized via the following loss:
\begin{align}
\mathcal{L}_{\Gx} = -\mathbb{E}_{x \sim p_{\text{data}}(x)}[\log \Dy(\Gx(x) | x)]\label{eq:Gx_loss}
\end{align}.
By constructing the adversarial relation, $\Gx$ and $\Dy$ are trained jointly and boost the other's performance gradually.

However, the desired style that $\Gx$ should learn is not specified in the aforementioned adversarial training, impacts the training stability and potentially results in model collapse.
Therefore, we introduce an extra loss term $\mathcal{L}_1$ that explicitly indicates the optimization direction:
\begin{equation}
    \mathcal{L}_{1} = \mathbb{E}_{(x, y) \sim p_{\text{data}}(x, y)}[\|y - \Gx(x)\|_1]\label{eq:L1_loss_xy}
\end{equation}.
While $\mathcal{L}_1$ guides the learning of the style, on the other hand, a too-low value hinders the injection of the necessary information for HPE.
Therefore, inspired by Huber loss \cite{huber1992robust}, we adopt a modified loss $\mathcal{L}_{\mathcal{X}\mathcal{Y}}$ to balance the style guidance and information injection:
\begin{equation}
    \mathcal{L}_{\mathcal{X}\mathcal{Y}} =
\begin{cases} 
    \mathcal{L}_{1} & \text{if } \mathcal{L}_{1} \geq T, \\
    0 & \text{otherwise}
\end{cases}\label{eq:L_xy}
\end{equation}
, where $T$ is a predefined threshold.
The total loss of the privacy-enhancing module is
\begin{equation}
    \mathcal{L}_{enhance} = \mathcal{L}_{\Dy} + \mathcal{L}_{\Gx} + \lambda_1 \mathcal{L}_{\mathcal{XY}}
\end{equation}
, where $\lambda_1$ is a hyperparameter.

The remaining background denoted $X_{n, \setminus} = X_n \setminus \{x_{n,0},\cdots, x_{n,i}\}$ is combined with the privacy-enhanced portraits $\{y^{\prime}_{n,0},\cdots, y^{\prime}_{n,i}\}$ to form the privacy-enhanced image $Y_n^\prime=X_{n,\setminus}\bigcup \{y^{\prime}_{n,0},\cdots, y^{\prime}_{n,i}\}$.

\subsection{Privacy Recovery Module}\label{sec:privacy_recovery_module}
The privacy recovery module aims to recover the SPI hidden in the privacy-enhanced portraits $y^\prime\in\mathcal{Y}$.
The recovery problem can be defined as follows:
Given the privacy-enhanced portraits $\{y_{0,0}^\prime,\cdots,y_{n,i}^\prime\}\in\mathcal{Y}$, the module recovers the SPI and generates the privacy-recovered portraits $\{x_{0,0}^\prime,\cdots,x_{n,i}^\prime\}\in\mathcal{X}$ as similar to the original portraits as possible.

The recovery module adopts a similar architecture as the privacy-enhancing module, consisting of a generator $\Gy$ and a discriminator $\Dx$.
However, one difference between the two modules is that the privacy recovery module takes the learnable generations $\{y_{n,i}^\prime,\cdots,y_{n,i}^\prime\}$ as input, but not the fixed inputs, such as $x_{n,i}$, and $y_{n,i}$.
This is because the goal of the recovery module is specific to recover the SPI in the privacy-enhanced portraits, therefore, there is no use in force it learns the mapping from the traditional desensitized images $y_{n,i}$ to $x_{n,i}$.
The generator $\Gy$ generates the privacy-recovered portrait $x_{n,i}^\prime=\Gy(y_{n,i}^\prime)$, and the discriminator $\Dx$ distinguishes the portrait pair $(x_{n,i}, x_{n,i}^\prime)$.
The $\Gy$ is optimized via the loss
\begin{align}
\mathcal{L}_{\Gy} = -\mathbb{E}_{y^\prime \sim p(y^\prime)}[\log \Dy(\Gy(y^\prime) | y^\prime)]\label{eq:Gy_loss}
\end{align}
, and the $\Dx$ facilitates its performance by the loss 
\begin{align}
    \mathcal{L}_{\Dx} = & -\mathbb{E}_{x\sim p_{\text{data}}(x), y^\prime\sim p(y^\prime)}[\log \Dx(x | y^\prime)] \notag \\
    & - \mathbb{E}_{y^\prime\sim p(y^\prime)}[\log(1 - \Dx(\Gy(y^\prime) | y^\prime))]\label{eq:Dx_loss}
\end{align}.
A consistency loss \cref{eq:consistency_loss} is introduced in the recovery module to guide the whole privacy-enhancing and recovery process explicitly.
It forces the recovered portraits to have a similar style to the original portraits.
\begin{equation}
    \mathcal{L}_{\text{consistency}} = \mathbb{E}_{x \sim p_{\text{data}}(x)}[\| \Gx(\Gy(x)) - x \|_1])\label{eq:consistency_loss}
\end{equation}.
The total objective function of the privacy recovery module is 
\begin{equation}
    \mathcal{L}_{recovery} = \mathcal{L}_{\Gy} + \mathcal{L}_{\Dx} + \lambda_2\mathcal{L}_{\text{consistency}}
\end{equation}
, where $\lambda_2$ is a hyperparameter that controls the style explicit guidance.

\subsection{Pose Estimator}\label{sec:pose_estimator}

The pose estimator model $\mathcal{P}$ conducts pose estimation on $Y_n^\prime$ without seeing any SPI. Given a set of images $\{Y_0^\prime,\cdots, Y_n^\prime\}$, the model is optimized via multiple loss terms: a bounding box loss $\mathcal{L}_{bbox}$ that measures the overlap between the predicted bounding box $[y_{n,i}^\prime]$ and the ground truth bounding box, a pose loss $\mathcal{L}_{pose}$ that measures the difference between the predicted keypoints and ground truth articulation keypoints, an object loss $\mathcal{L}_{obj}$ that classifies whether a keypoint is visible, and a classification loss $\mathcal{L}_{cls}$ that classifies the detected objects into predefined category (i.e., ``human'').
The loss function of a pose estimator is
\begin{equation}
\mathcal{L}_{PE_{\mathcal{Y}}}=\mathcal{L}_{bbox_{\mathcal{Y}}}+\mathcal{L}_{pose_{\mathcal{Y}}}+\mathcal{L}_{obj_{\mathcal{Y}}}+\mathcal{L}_{cls_{\mathcal{Y}}}\label{eq:pe_y_loss}
\end{equation}.
Since the purpose of our system is to estimate human pose in both the privacy-enhanced images and the privacy-recovered images, $\mathcal{P}$ should be capable of implementing pose estimation on the images from both domains ($\mathcal{X}$ and $\mathcal{Y}$).
Therefore, the pose estimator is trained on the pairs $(y^\prime, x^\prime)$.
The total loss for the pose estimator is denoted as:
\begin{equation}
    \mathcal{L}_{PE} = \mathcal{L}_{PE_{\mathcal{X}}} + \mathcal{L}_{PE_{\mathcal{Y}}}\label{eq:pe_loss}
\end{equation}.
$\mathcal{L}_{PE_{\mathcal{X}}}$ is defined on recovered images and $\mathcal{L}_{PE_{\mathcal{Y}}}$ is for privacy-enhanced images. 

Finally, we jointly optimize the privacy-enhancing, privacy-recovery, and pose estimation modules end-to-end with the following overall loss function:
\begin{equation}
\mathcal{L}=\mathcal{L}_{enhance}+\mathcal{L}_{recovery}+\lambda_3\mathcal{L}_{PE}\label{eq:total_loss}
\end{equation}
, where $\lambda_3$ is a hyperparameter.

\begin{figure*}[t]
  \centering
   \includegraphics[width=0.9\linewidth]{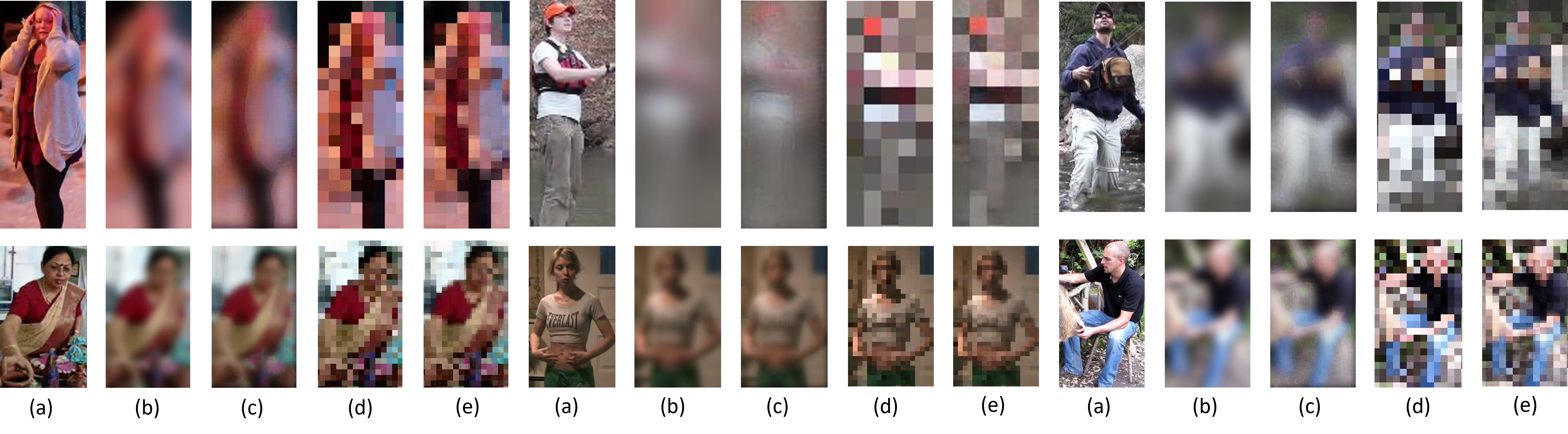}

   \caption{Qualitative comparison on privacy-enhanced portraits. (a) original portraits; (b)/(d) conventional desensitized portraits via blurring/pixelation;
(c)/(e) privacy-enhanced portraits guided by blurring/pixelation. Enlarge for details.}
   \label{fig:privacy enhancement}
      \vspace{-1em}
\end{figure*}

\begin{table*}[]
\centering
\resizebox{0.99\textwidth}{!}{%
\begin{threeparttable}
\begin{tabular}{ccccccccc}
\toprule
\multicolumn{1}{c|}{\textbf{Dataset}} & \multicolumn{4}{c|}{\textbf{MPII} ($\text{mAP@0.5}_{\{\text{pre, o}\}} = 83.9$, $\text{mAR@0.5}_{\{\text{pre, o}\}} = 89.4$)\tnote{a}}                                                               & \multicolumn{4}{c}{\textbf{COCO} ($\text{mAP@0.5}_{\{\text{pre, o}\}} = 86.2$, $\text{mAR@0.5}_{\{\text{pre, o}\}} = 90.8$)\tnote{a}}                                           \\ \midrule
\multicolumn{1}{c|}{\textbf{Metrics}} & PSNR(o,p)$\downarrow$\tnote{b} & SSIM(o,p)$\downarrow$\tnote{b} & $\text{mAP@0.5}_{\{\text{pre, p}\}}\downarrow$\tnote{c} & \multicolumn{1}{c|}{$\text{mAR@0.5}_{\{\text{pre, p}\}}\downarrow$\tnote{c}} & PSNR(o,p)$\downarrow$\tnote{b} & SSIM(o,p)$\downarrow$\tnote{b} & $\text{mAP@0.5}_{\{\text{pre, p}\}}\downarrow$\tnote{c} & $\text{mAR@0.5}_{\{\text{pre, p}\}}\downarrow$\tnote{c} \\ \midrule
\multicolumn{9}{l}{\textbf{(1). Blurring}}                                                                                                                                                           \\ \midrule
\multicolumn{1}{c|}{Conventional}     &    23.71       &    0.65       &      0.3     &       1.0                                     &  23.01         &   0.60        &      27.2       &   31.4                   \\
\multicolumn{1}{c|}{Ours}          &   23.36        &   0.68        &      11.9       &       18.7                                    &  22.81         &  0.66         &  35.3          &  40.5                    \\ \midrule
\multicolumn{9}{l}{\textbf{(2). Pixelation}}                                                                                                                                                         \\ \midrule
\multicolumn{1}{c|}{Conventional}     &   19.97        &  0.53         &    0.1       &   0.6                                        &  19.34         & 0.49          & 0.1           & 0.3                     \\
\multicolumn{1}{c|}{Ours}       &   20.89      &   0.56    &  0.2   &  0.5                             &20.15           &0.54           &0.2          &0.3                      \\ \bottomrule
\end{tabular}%
\begin{tablenotes}[leftmargin=0.35cm, itemindent=.0cm, itemsep=0.0cm, topsep=0.1cm]
\item[a] 
The subscript \{pre, o\} indicates a pose estimator pretrained on original images (pre), and tested on original images (o).
\item[b] A lower value indicates a lower similarity between the original image (o) and the privacy-enhanced image (p), showing a better privacy enhancement.
\item[c]
A lower value indicates a better privacy enhancement. The subscript \{pre, p\} represents a pose estimator pretrained on original images (pre), and tested on privacy-enhanced images (p).
\end{tablenotes}
\end{threeparttable}
}
\caption{Image Quality and Pose Estimation Performance of Privacy-enhanced Portraits.}
\label{tab:privacy enhancement}
   \vspace{-1em}
\end{table*}

\section{Experiments}\label{sec:experiments}

\subsection{Setup}
Our system is developed using PyTorch \cite{paszke2019pytorch} and is trained on an NVIDIA RTX A6000 GPU. 
The architecture employs a U-Net \cite{ronneberger2015u} model as the backbone for the generators and PatchGAN \cite{isola2017image} for the discriminators. 
For pose estimation, we integrate YOLOv8 \cite{Jocher_Ultralytics_YOLO_2023}, although the model can be interchangeably replaced with alternative pose estimation algorithms to suit specific needs. 
Training of these modules employs distinct optimization strategies: the generators and discriminators utilize the Adam optimizer, whereas YOLOv8 employs the AdamW optimizer to potentially enhance training stability and performance. 
The initial learning rate is set at $0.000035$, which undergoes exponential decay to facilitate convergence. 
Data augmentation techniques include random horizontal flipping and adjustments to hue, saturation, and brightness of the input images. We train our models with a batch size of 16.

The experiments are conducted on the widely used datasets: MPII Human Pose (MPII) \cite{andriluka14cvpr}, 
and Microsoft Common Objects in Context (COCO) \cite{lin2014microsoft}. 
The MPII dataset comprises approximately 25,000 images featuring over 40,000 individuals.
Each pose within this dataset is manually annotated with up to 16 body joints. 
The COCO dataset encompasses over 200,000 labeled individuals, each annotated with 17 body joints, primarily focusing on people depicted at medium and large scales.

We assess our system utilizing established metrics for image quality and pose estimation. 
For the evaluation of privacy enhancement and recovery, we employ two commonly accepted metrics: the Peak Signal-to-Noise Ratio (PSNR) and the Structural Similarity Index Measure (SSIM) \cite{wang2004image}. 
PSNR values range from 0 to $\infty$, with $\infty$ indicating perfect similarity, implying no discernible difference between the compared images.
The SSIM varies from 0 to 1, where a value of 0 indicates no structural similarity between the images.
Typically, image pairs are deemed to exhibit high similarity when the PSNR$\geq 30$ and the SSIM $\geq 0.9$ \cite{wang2004image, huynh2008scope}.
For pose estimation, we utilize the Object Keypoint Similarity (OKS), analogous to the Intersection over Union (IoU) used in object detection. 
OKS is calculated based on the scale of the subject and the Euclidean distances between predicted keypoints and their corresponding ground truth points. 
To quantify the performance of our pose estimation, we employ the mean Average Precision (mAP) and mean Average Recall (mAR) at an OKS threshold of 0.5, denoted as mAP@0.5 and mAR@0.5.

The conventional desensitization methods used in our system comprise Gaussian blurring, where the kernel radius $r$ is set to 8, and pixelation, with each pixel block having a side length of $r=12$.

\begin{figure*}[t]
  \centering
   \includegraphics[width=0.7\textwidth]{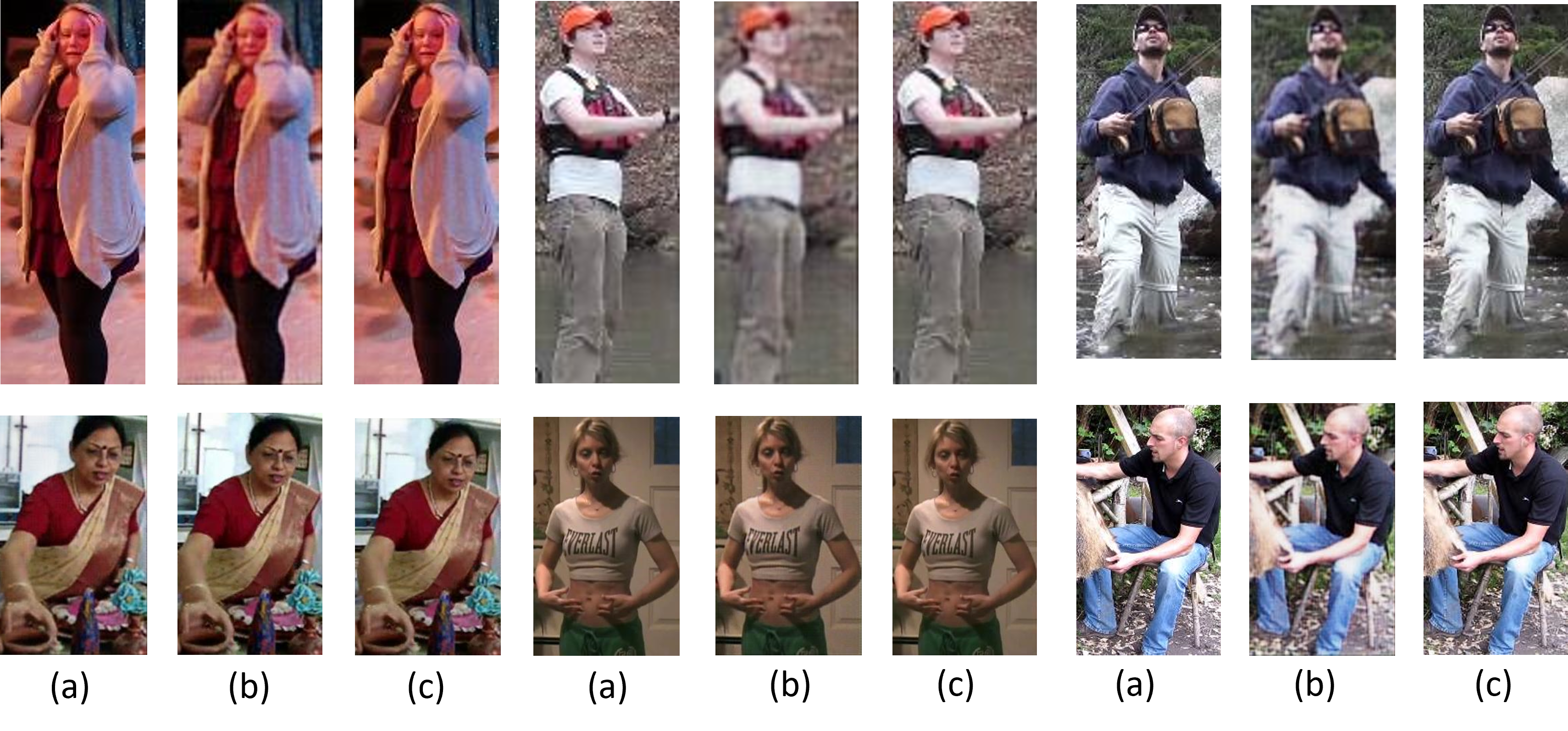}
   \caption{Qualitative results of the privacy-recovered portraits. (a) original portraits; (b)/(c) the portraits recovered from the privacy-enhanced portraits guided by blurring/pixelation. Enlarge for details.}
   \label{fig:privacy recovery}
\end{figure*}

\begin{table*}[t]
\centering
\resizebox{0.9\textwidth}{!}{%
\begin{threeparttable}
\begin{tabular}{ccccccc}
\toprule
\multicolumn{1}{c|}{\textbf{Dataset}} &
  \multicolumn{6}{c}{MPII ($\text{mAP@0.5}_{\{\text{pre, o}\}} = 83.9$, $\text{mAR@0.5}_{\{\text{pre, o}\}} = 89.4$)\tnote{a}} \\ \midrule
\multicolumn{1}{c|}{\textbf{Metrics}} &
  $\text{mAP@0.5}_{\{\text{joint, p}\}}\uparrow$\tnote{b} &
  $\text{mAR@0.5}_{\{\text{joint, p}\}}\uparrow$\tnote{b} &
  $\text{mAP@0.5}_{\{\text{joint, r}\}}\uparrow$\tnote{b} &
  $\text{mAR@0.5}_{\{\text{joint, r}\}}\uparrow$\tnote{b} &
  PSNR(o,r)$\uparrow$\tnote{c} &
  SSIM(o,r)$\uparrow$\tnote{c} \\ \hline
\multicolumn{7}{l}{\textbf{(1). Blurring}} \\ \hline
\multicolumn{1}{c|}{Conventional} &
  70.5 &
  81.3 &
  - &
  - &
  - &
  - \\
\multicolumn{1}{c|}{Ours} &
  \textbf{81.5} &
  \textbf{88.8} &
  \textbf{87.4} &
  \textbf{92.4} &
  32.58 &
  0.94 \\ \hline
\multicolumn{7}{l}{\textbf{(2). Pixelation}} \\ \hline
\multicolumn{1}{c|}{Conventional} &
  65.2 &
  77.9 &
  - &
  - &
  - &
  - \\
\multicolumn{1}{c|}{Ours} &
  74.9 &
  84.9 &
  87.1 &
  91.3 &
  \textbf{38.54} &
  \textbf{0.98} \\ \midrule
\multicolumn{1}{c|}{\textbf{Dataset}} &
  \multicolumn{6}{c}{COCO ($\text{mAP@0.5}_{\{\text{pre, o}\}} = 86.2$, $\text{mAR@0.5}_{\{\text{pre, o}\}} = 90.8$)\tnote{a}} \\ \midrule
\multicolumn{1}{c|}{\textbf{Metrics}} &
  $\text{mAP@0.5}_{\{\text{joint, p}\}}\uparrow$\tnote{b} &
  $\text{mAR@0.5}_{\{\text{joint, p}\}}\uparrow$\tnote{b} &
  $\text{mAP@0.5}_{\{\text{joint, r}\}}\uparrow$\tnote{b} &
  $\text{mAR@0.5}_{\{\text{joint, r}\}}\uparrow$\tnote{b} &
  PSNR(o,r)$\uparrow$\tnote{c} &
  SSIM(o,r)$\uparrow$\tnote{c} \\ \hline
\multicolumn{7}{l}{\textbf{(1). Blurring}} \\ \hline
\multicolumn{1}{c|}{Conventional} &
  62.1 & 
  74.7 & 
  - & 
  - &
  - &
  - \\
\multicolumn{1}{c|}{Ours} & \textbf{75.3}
   & \textbf{84.9}
   & \textbf{89.0}
   & \textbf{92.5}
   &34.92
   &0.95
   \\ \hline
\multicolumn{7}{l}{\textbf{(2). Pixelation}} \\ \hline
\multicolumn{1}{c|}{Conventional} &
  59.4 & 
  65.6 & 
  - &
  - &
  - &
  - \\
\multicolumn{1}{c|}{Ours} & 70.3
   & 81.1
   & 88.6
   & 92.0
   &\textbf{37.63}
   &\textbf{0.98}
   \\ \bottomrule
\end{tabular}
\vspace{0.5em}
\begin{tablenotes}[leftmargin=0.35cm, itemindent=.0cm, itemsep=0.0cm, topsep=0.1cm]
\item[a] The subscript \{pre, o\} indicates a pose estimator pretrained on original images (pre), and tested on original images (o).
\item[b] 
A higher value indicates a better performance. The subscript $\{\text{joint, o}\}, \{\text{joint, p}\}, \{\text{joint, r}\}$ represents a pose estimator joint trained with the privacy-enhancing and recovery modules, and tested on the original images (o), privacy-enhanced images (p), and privacy recovery images (r), respectively.
\item[c] 
A higher value indicates a higher similarity between the original image (o) and the privacy-recovered image (r), showing a better recovery.
\end{tablenotes}
\end{threeparttable}
}
\vspace{-1em}
\caption{Image Quality and Pose Estimation Performance of Privacy-recovered Portraits.}
\label{tab:privacy recovery}
\vspace{-1.5em}
\end{table*}

\subsection{Results of Privacy Enhancement}\label{sec: results of privacy enhancement}
\Cref{fig:privacy enhancement} presents a qualitative comparison of our privacy-enhancing module. 
In contrast to the original portraits, our privacy-enhanced portraits demonstrate superior visual privacy protection. 
The contours of the body, as well as the details of the face and clothing, are obscured, thereby preventing SPI through visual inspection. 
Compared to conventional desensitized portraits, our privacy-enhanced portraits achieve a competitive level of visual obfuscation while employing a distinct learnable approach.

\Cref{tab:privacy enhancement} illustrates the quantitative comparison between privacy-enhanced portraits and raw images in terms of PSNR and SSIM. We also show the zero-shot HPE performance of a pretrained pose estimator pre-trained on both types of privacy-enhanced images. 
Compared to conventional desensitized portraits, our privacy-enhanced portraits attain similar PSNR and SSIM values when compared to the original portraits. This indicates that our method achieves comparable levels of privacy protection to the baseline.
Effective privacy enhancement necessitates that the pose estimator, pretrained on original images, should fail to make accurate zero-shot inferences on the privacy-enhanced images. 
Our method significantly reduces the HPE performance, indicating that the pretrained pose estimator struggles to perform HPE accurately on the modified images. 
This substantial degradation in performance demonstrates the robust privacy enhancement capabilities of our approach.
The privacy-enhancing module guided by pixelation demonstrates a lower image similarity to the original images and more significantly impacts the HPE performance of the pretrained pose estimator, compared to the module guided by blurring.

In addition to the visual obfuscation capability, we also expect the system to restore high efficacy in HPE by finetuning the pose estimator and adapting toward the privacy-enhanced images. However, with the conventional method, the carefully finetuned pose estimator model still observes a significant drop in performance. As shown in \cref{tab:privacy recovery}, the metric $\text{mAP@0.5}_{\{\text{joint, p}\}}$ was cut by around 15\%, mainly due to the irrecoverable loss of information with the obfuscation.
In contrast, when enabling the joint optimization of the three components within our system, there is a significant improvement in HPE performance, as evidenced by the data in the first two columns of \cref{tab:privacy recovery}.
Both the mAP@0.5 and the mAR@0.5 of our method exceed those achieved with conventional desensitized portraits, with about 10\% improvement in mAP\@0.5.
Although these values are still marginally lower than those obtained by applying a pose estimator trained on original images to original images, this underscores that our system effectively incorporates valuable information into the privacy-enhanced portraits, thereby enhancing HPE performance.

\subsection{Results of Privacy Recovery} \label{sec: results of privacy recovery}

A key strength of our system is that the anonymization process is reversible and we learn a uniform pose estimator for images before and after recovery. 
\Cref{fig:privacy recovery} provides a qualitative comparison between the original portraits and the privacy-recovered portraits.
The privacy-recovered portraits display visual quality that is on par with the original portraits. 
Distinguishing between the original and the privacy-recovered portraits through human visual inspection proves to be challenging, indicating effective restoration of SPI in the privacy-recovered images.

\Cref{tab:privacy recovery} presents the image quality metrics for the recovered images. 
The PSNR and SSIM values of the privacy-recovered portraits relative to the original portraits (i.e., PSNR(o, r) and SSIM(o,r) in \cref{tab:privacy recovery}) exceed 30 and 0.9, respectively, demonstrating that the privacy recovery module effectively restores the SPI.
Surprisingly, the pose estimator, optimized jointly with other system components, outperforms a pose estimator trained solely on original images when applied to those images; it shows an average 3\% improvement in mAP.
This improvement is likely due to the privacy recovery module's dual function of not only restoring SPI from the portraits but also enhancing the HPE-related features during the recovery process, as guided by  $\mathcal{L}_{PE_{\mathcal{X}}}$. 
Consequently, the privacy-recovered portraits retain the SPI while accentuating HPE-related features, thereby facilitating more accurate pose estimation.
Additionally, the experimental results show that the system guided by blurring outperforms the other one (i.e., guided by pixelation) in terms of pose estimation on obfuscated and recovered images. 
Conversely, the system guided by pixelation more effectively restores the SPI from the privacy-enhanced images, achieving higher image quality (i.e., PSNR and SSIM metrics).

\begin{table}[]
\setlength{\tabcolsep}{1pt}
\centering
\resizebox{0.95\columnwidth}{!}{%
\begin{threeparttable}
\begin{tabular}{ccccc}
\toprule
\multicolumn{1}{c|}{\textbf{Metrics}} &
  $\text{PSNR(o, p)}\downarrow$ &
  $\text{SSIM(o, p)}\downarrow$ &
  $\text{mAP@0.5}_{\{\text{joint, p}\}}\uparrow$ &
  $\text{mAR@0.5}_{\{\text{joint, p}\}}\uparrow$ \\ \midrule
\multicolumn{5}{l}{\textbf{(1). Blurring}\tnote{a}}   \\ \midrule
\multicolumn{1}{c|}{$r = 2$}         & 29.22        & 0.84      & 82.5      & 89.3       \\
\multicolumn{1}{c|}{$r = 4$}         & 26.45        & 0.73      & 82.1      & 89.0       \\
\multicolumn{1}{c|}{$r = 8$}         & 23.36        & 0.68      & 81.5      & 88.8       \\
\multicolumn{1}{c|}{$r = 12$}        & 22.49        & 0.63      & 77.8      & 86.3       \\ \midrule
\multicolumn{5}{l}{\textbf{(2). Pixelation}\tnote{b}} \\ \midrule
\multicolumn{1}{c|}{$r = 4$}         & 24.40        & 0.68      & 80.8      & 88.2      \\
\multicolumn{1}{c|}{$r = 8$}         & 21.36        & 0.55      & 76.2      & 85.3      \\
\multicolumn{1}{c|}{$r = 12$}        & 20.89        & 0.56      & 74.9      & 84.9      \\
\multicolumn{1}{c|}{$r = 16$}        & 20.09        & 0.47      & 60.7      & 75.2      \\ \bottomrule
\end{tabular}
\begin{tablenotes}
\item[a] In blurring, $r$ represents the radius of the blur kernel.
\item[b] In pixelation, $r$ denotes the side length of each pixel block.
\end{tablenotes}
\end{threeparttable}

}
\caption{Impact of Conventional Desensitization Guidance.}
\label{tab:impact on conv desens}
\end{table}

\section{Discussion}\label{sec:discussion}

\subsection{Impact of 
Desensitization Guidance}\label{sec:desensitization}

Conventional desensitization guidance dictates the level of privacy enhancement in our module, influencing the style of the generated privacy-enhanced portraits. 
Severe desensitization, while increasing privacy, complicates the integration of HPE-related features, thereby hindering the joint training of the pose estimator and adversely affecting HPE performance. 
Conversely, mild desensitization facilitates feature integration but may compromise privacy enhancement. 
Thus, the strategic selection of desensitization levels is crucial, as it significantly impacts overall system performance.
\Cref{tab:impact on conv desens} presents the performance of various conventional desensitization guidance methods, evaluating both portrait quality and HPE accuracy.
As $r$ increases, the capability for privacy enhancement improves, whereas the HPE performance deteriorates. Specifically, when $r$ increases from 12 to 16 in pixelation, the similarity between the privacy-enhanced portraits and the original portraits remains relatively unchanged, yet there is a substantial decline in HPE performance. 
A similar pattern is observed in blurring when $r$ changes from 8 to 12.

\subsection{Impact of Adopting Noise Addition as Privacy Enhancement Guidance}\label{sec:noise}
\begin{figure}[t]
  \centering
   \includegraphics[width=0.95\columnwidth]{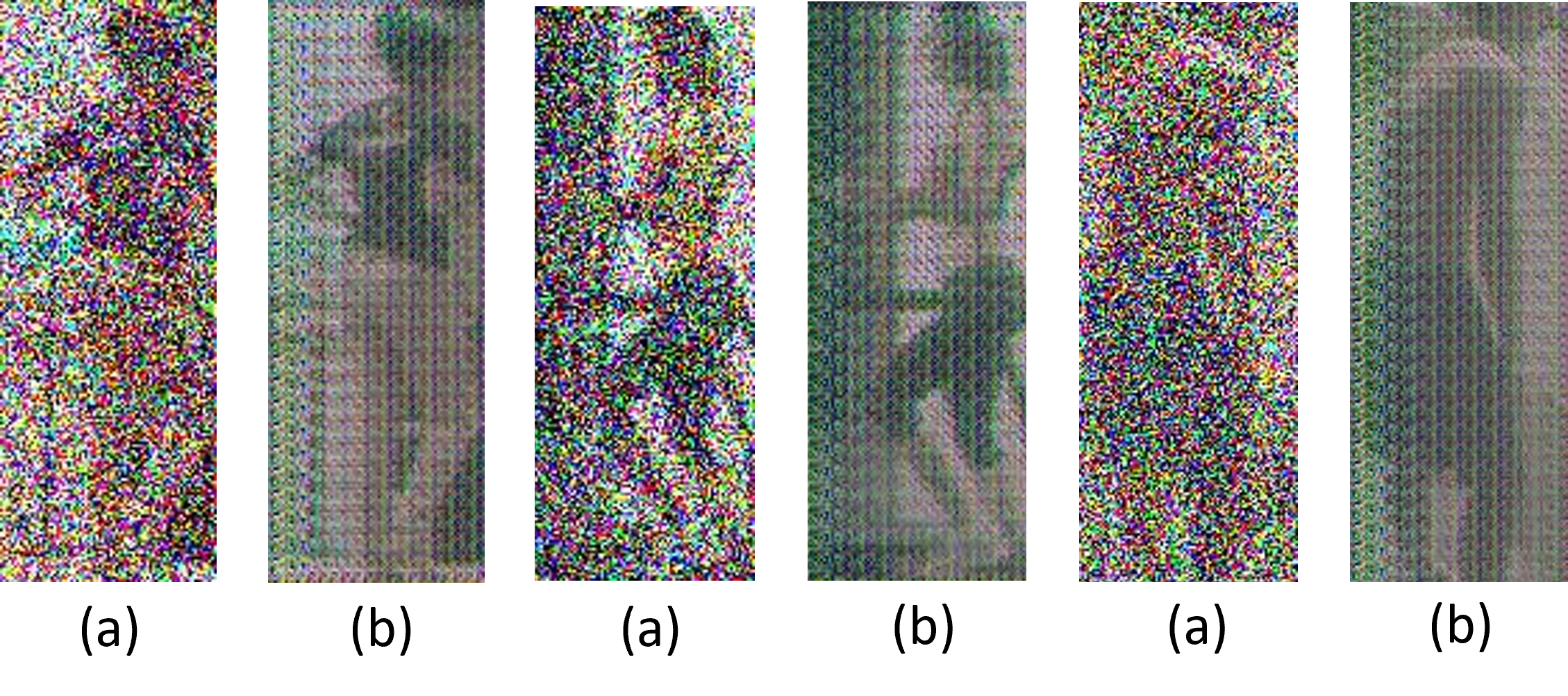}
   \vspace{-1em}
   \caption{Qualitative results of the privacy-enhanced portraits guided by Gaussian noise. (a) conventional desensitized portraits; (b) privacy-enhanced portraits guided by Gaussian noise addition.}
   \label{fig:noise}
   \vspace{-1em}
\end{figure}
Gaussian noise addition is another widely recognized conventional desensitization method. 
We explore its impact when utilized as guidance within our system. 
\Cref{fig:noise} presents a qualitative comparison between conventional desensitized portraits and their corresponding privacy-enhanced counterparts. 
The privacy-enhanced portraits generated in the system exhibit numerous artifacts, diverging from the guaidance and compromising privacy preservation. 
We hypothesize that this deviation arises because Gaussian noise addition introduces a random pattern, which is challenging to learn through \cref{eq:L1_loss_xy}.

\subsection{Backbone \& Model Lightweightness}\label{sec:lightweightness}
To facilitate deployment in surveillance environments, our privacy-enhancing module must be sufficiently lightweight to operate on edge devices without sacrificing its privacy-enhancement capabilities. 
We evaluate the impact of different backbones of the privacy-enhancing module on overall performance.
\Cref{tab:impact of backbone} displays the results in terms of privacy-enhancement, HPE performance, and inference speed.
Although both U-Net and ResNet backbones effectively capture the patterns of conventional desensitization, the privacy-enhanced portraits generated with a ResNet backbone exhibit poorer HPE performance.
This suggests a failure in integrating HPE-related features effectively, potentially due to the absence of skip connections that are present in U-Net for transferring low-level information across the network.
Further evaluations conducted on the NVIDIA Jetson AGX Orin \cite{NvidiaJetsonOrin2023} reveal that U-Net configurations 7 and 8 achieve desirable inference speeds, maintaining real-time processing capabilities (i.e., 30 FPS), which surpass those of the ResNet backbones. 
Given these findings, U-Net emerges as the more suitable backbone for our privacy-enhancing module, considering both performance metrics and latency requirements.

\begin{table}[]
\setlength{\tabcolsep}{1pt}
\resizebox{1\columnwidth}{!}{%
\begin{threeparttable}
\begin{tabular}{cccccccc}
\toprule
\multicolumn{1}{c|}{\textbf{Metrics}} &
  \multicolumn{1}{c}{PSNR(o, p)$\downarrow$} &
  \multicolumn{1}{c}{SSIM(o, p)$\downarrow$} &
  \multicolumn{1}{c}{PSNR(o, r)$\uparrow$} &
  \multicolumn{1}{c}{SSIM(o, r)$\uparrow$} &
  $\text{mAP@0.5}_{\{\text{joint, p}\}}\uparrow$ &
  $\text{mAR@0.5}_{\{\text{joint, p}\}}\uparrow$ &
  FPS$\uparrow$\tnote{a} \\ \midrule
\multicolumn{8}{l}{\textbf{(1). Blurring}}         \\ \midrule
\multicolumn{1}{c|}{U-Net 7}  &23.36  &0.68  &32.58  &0.94  &81.5  &88.8  &63.71 \\
\multicolumn{1}{c|}{U-Net 8}  &23.34  &0.68  &32.55  &0.94  &81.4  &88.8  &59.84 \\
\multicolumn{1}{c|}{ResNet 6} &23.86  &0.69  &32.46  &0.93  &68.7  &79.1  &39.95 \\
\multicolumn{1}{c|}{ResNet 9} &23.91  &0.69  &32.41  &0.93  &67.5  &78.8  &34.82 \\ \midrule
\multicolumn{8}{l}{\textbf{(2). Pixelation}}           \\ \midrule
\multicolumn{1}{c|}{U-Net 7}  &20.89  &0.56  &38.54  &0.98  &74.9  &84.9  &61.22 \\
\multicolumn{1}{c|}{U-Net 8}  &20.81  &0.55  &38.63  &0.98  &75.2  &85.0  &57.19\\
\multicolumn{1}{c|}{ResNet 6} &21.15  &0.57  &38.51  &0.98  &60.5  &73.9  &37.46 \\
\multicolumn{1}{c|}{ResNet 9} &21.18  &0.57  &38.45  &0.97  &60.9  &74.1  &33.96 \\ \bottomrule
\end{tabular}%
\begin{tablenotes}[leftmargin=0.35cm, itemindent=.0cm, itemsep=0.0cm, topsep=0.1cm]
\item[a] Frame per second (FPS) is measured at inference speed on NVIDIA Jetson AGX Orin.
\end{tablenotes}
\end{threeparttable}
}
\vspace{-0.5em}
\caption{Impact of Backbone Architecture and Inference Speed on Edge Device.}
\label{tab:impact of backbone}
   \vspace{-1em}
\end{table}

\section{Related Work}\label{sec:related}
\subsection{Pose Estimation}
Multiple approaches exist for addressing HPE, with recent advancements in deep learning demonstrating superior performance compared to earlier methods
\cite{pishchulin2013poselet, yang2011articulated, yao2010modeling, wang2013beyond}. 
Notable recent deep-learning-based algorithms include \cite{li2021pose, mao2022poseur, artacho2021omnipose, geng2023human, li2021tokenpose}.
These methods are typically discussed separately concerning single-person and multi-person scenarios. 
In single-person pose estimation, the objective is to localize joint positions in images containing only one person \cite{mao2022poseur, toshev2014deeppose,chen2014articulated, tompson2015efficient}. 
In contrast, multi-person pose estimation methods can be categorized into top-down and bottom-up approaches. Top-down methods \cite{cai2020learning, wang2020graph, chen2018cascaded, li2021tokenpose, shi2022end, yang2021transpose} first employ person detectors to identify individual persons in an image, then apply single-person pose estimation to each detected person. In contrast, bottom-up methods \cite{wang2022lite, wang2022regularizing, luo2021rethinking} first detect all body keypoints in an image and subsequently group them into distinct person instances.

\subsection{Privacy Enhancing Methods}
Naive image privacy-enhancing techniques such as masking, blurring, or pixelation are commonly employed in practice \cite{agrawal2011person, chen2007tools}. However, these methods tend to remove semantic information and significantly degrade the quality of privacy-enhanced images, rendering the data unusable for many applications. Some efforts have explored addressing the issue through additional modalities \cite{paolanti2018person, ahmad2023person}, but these approaches are often impractical and lack scalability.
\cite{zhao2022freed, cheng2019person} involve encrypting feature vectors of visual data to ensure privacy. However, encrypting large volumes of visual data is complex and resource-intensive.
Recent studies have leveraged deep generative models to anonymize data while preserving its utility for downstream applications.
They either inpaint missing regions \cite{hukkelaas2019deepprivacy, maximov2020ciagan} or transform original regions \cite{gafni2019live, ren2018learning, lopez2024privacy, xiao2020adversarial}. 
However, much of prior work has focused primarily on face anonymization, leaving other identifiers such as clothing and body type untouched, which can compromise privacy.
While some efforts have targeted full-body anonymization \cite{hukkelaas2023deepprivacy2, maximov2020ciagan}, these approaches often lack recoverability, limiting their applicability. 

\section{Conclusion}\label{sec:conclusion}
We propose a privacy-enhancing system for HPE that addresses the critical need for protecting SPI while maintaining the performance of HPE tasks. 
The privacy-enhancing module, privacy recovery module, and pose estimator work in unison to anonymize SPI, allow for its recovery by authorized personnel, and ensure the preservation of contextual information essential for accurate behavior interpretation.
Our experimental results demonstrate that the system achieves robust performance in privacy protection, accurate recovery of original images, and high-precision HPE. 
{\small
\bibliographystyle{ieee_fullname}
\bibliography{egbib}

\begin{thebibliography}{10}\itemsep=-1pt

\bibitem{agrawal2011person}
Prachi Agrawal et~al.
\newblock Person de-identification in videos.
\newblock {\em IEEE Transactions on Circuits and Systems for Video Technology}, 21(3):299--310, 2011.

\bibitem{ahmad2023person}
Shafiq Ahmad et~al.
\newblock Person re-identification without identification via event anonymization.
\newblock In {\em Proceedings of the IEEE/CVF International Conference on Computer Vision}, pages 11132--11141, 2023.

\bibitem{ahmad2024event}
Shafiq Ahmad et~al.
\newblock Event anonymization: Privacy-preserving person re-identification and pose estimation in event-based vision.
\newblock {\em IEEE Access}, 2024.

\bibitem{andriluka14cvpr}
Mykhaylo Andriluka et~al.
\newblock 2d human pose estimation: New benchmark and state of the art analysis.
\newblock In {\em IEEE Conference on Computer Vision and Pattern Recognition (CVPR)}, June 2014.

\bibitem{artacho2021omnipose}
Bruno Artacho et~al.
\newblock Omnipose: A multi-scale framework for multi-person pose estimation.
\newblock {\em arXiv preprint arXiv:2103.10180}, 2021.

\bibitem{barnoviciu2019gdpr}
Eduard Barnoviciu et~al.
\newblock Gdpr compliance in video surveillance and video processing application.
\newblock In {\em 2019 International Conference on Speech Technology and Human-Computer Dialogue (SpeD)}, pages 1--6. IEEE, 2019.

\bibitem{cai2020learning}
Yuanhao Cai et~al.
\newblock Learning delicate local representations for multi-person pose estimation.
\newblock In {\em Computer Vision--ECCV 2020: 16th European Conference, Glasgow, UK, August 23--28, 2020, Proceedings, Part III 16}, pages 455--472. Springer, 2020.

\bibitem{cao2022bed}
Ting Cao et~al.
\newblock In-bed human pose estimation from unseen and privacy-preserving image domains.
\newblock In {\em 2022 IEEE 19th International Symposium on Biomedical Imaging (ISBI)}, pages 1--5. IEEE, 2022.

\bibitem{chadha2021deepfake}
Anupama Chadha et~al.
\newblock Deepfake: an overview.
\newblock In {\em Proceedings of second international conference on computing, communications, and cyber-security: IC4S 2020}, pages 557--566. Springer, 2021.

\bibitem{chen2007tools}
Datong Chen et~al.
\newblock Tools for protecting the privacy of specific individuals in video.
\newblock {\em EURASIP Journal on Advances in Signal Processing}, 2007:1--9, 2007.

\bibitem{chen2024taskclip}
Hanning Chen et~al.
\newblock Taskclip: Extend large vision-language model for task oriented object detection.
\newblock {\em arXiv preprint arXiv:2403.08108}, 2024.

\bibitem{chen2014articulated}
Xianjie Chen et~al.
\newblock Articulated pose estimation by a graphical model with image dependent pairwise relations.
\newblock {\em Advances in neural information processing systems}, 27, 2014.

\bibitem{chen2018cascaded}
Yilun Chen et~al.
\newblock Cascaded pyramid network for multi-person pose estimation.
\newblock In {\em Proceedings of the IEEE conference on computer vision and pattern recognition}, pages 7103--7112, 2018.

\bibitem{cheng2019person}
Hang Cheng et~al.
\newblock Person re-identification over encrypted outsourced surveillance videos.
\newblock {\em IEEE Transactions on Dependable and Secure Computing}, 18(3):1456--1473, 2019.

\bibitem{chik2013singapore}
Warren~B Chik.
\newblock The singapore personal data protection act and an assessment of future trends in data privacy reform.
\newblock {\em Computer Law \& Security Review}, 29(5):554--575, 2013.

\bibitem{dave2022spact}
Ishan~Rajendrakumar Dave et~al.
\newblock Spact: Self-supervised privacy preservation for action recognition.
\newblock In {\em Proceedings of the IEEE/CVF Conference on Computer Vision and Pattern Recognition}, pages 20164--20173, 2022.

\bibitem{dressel2021dangers}
Julia Dressel et~al.
\newblock The dangers of risk prediction in the criminal justice system.
\newblock 2021.

\bibitem{ebadinezhad2023systematic}
Sahar Ebadinezhad.
\newblock A systematic literature review on information security leakage: Evaluating security threat.
\newblock In {\em Proceedings of Third International Conference on Sustainable Expert Systems: ICSES 2022}, pages 993--1007. Springer, 2023.

\bibitem{fiske2020social}
Susan T~Tufts Fiske.
\newblock Social cognition: From brains to culture.
\newblock 2020.

\bibitem{gafni2019live}
Oran Gafni et~al.
\newblock Live face de-identification in video.
\newblock In {\em Proceedings of the IEEE/CVF International Conference on Computer Vision}, pages 9378--9387, 2019.

\bibitem{garcia2023human}
Guillermo Garcia-Cobo et~al.
\newblock Human skeletons and change detection for efficient violence detection in surveillance videos.
\newblock {\em Computer Vision and Image Understanding}, 233:103739, 2023.

\bibitem{geng2023human}
Zigang Geng et~al.
\newblock Human pose as compositional tokens.
\newblock In {\em Proceedings of the IEEE/CVF Conference on Computer Vision and Pattern Recognition}, pages 660--671, 2023.

\bibitem{gifford2007environmental}
Robert Gifford.
\newblock Environmental psychology: Principles and practice.
\newblock 2007.

\bibitem{hinojosa2021learning}
Carlos Hinojosa et~al.
\newblock Learning privacy-preserving optics for human pose estimation.
\newblock In {\em Proceedings of the IEEE/CVF international conference on computer vision}, pages 2573--2582, 2021.

\bibitem{huang2021exploration}
Wenjun Huang et~al.
\newblock Exploration of using a pressure sensitive mat for respiration rate and heart rate estimation.
\newblock In {\em 2021 43rd Annual International Conference of the IEEE Engineering in Medicine \& Biology Society (EMBC)}, pages 298--301. IEEE, 2021.

\bibitem{huang2024ecosense}
Wenjun Huang et~al.
\newblock Ecosense: Energy-efficient intelligent sensing for in-shore ship detection through edge-cloud collaboration.
\newblock {\em arXiv preprint arXiv:2403.14027}, 2024.

\bibitem{huang2024intelligent}
Wenjun Huang et~al.
\newblock Intelligent sensing framework: Near-sensor machine learning for efficient data transmission.
\newblock {\em IEEE Sensors Journal}, 2024.

\bibitem{huber1992robust}
Peter~J Huber.
\newblock Robust estimation of a location parameter.
\newblock In {\em Breakthroughs in statistics: Methodology and distribution}, pages 492--518. Springer, 1992.

\bibitem{hukkelaas2019deepprivacy}
H{\aa}kon Hukkel{\aa}s et~al.
\newblock Deepprivacy: A generative adversarial network for face anonymization.
\newblock In {\em International symposium on visual computing}, pages 565--578. Springer, 2019.

\bibitem{hukkelaas2023deepprivacy2}
H{\aa}kon Hukkel{\aa}s et~al.
\newblock Deepprivacy2: Towards realistic full-body anonymization.
\newblock In {\em Proceedings of the IEEE/CVF winter conference on applications of computer vision}, pages 1329--1338, 2023.

\bibitem{huynh2008scope}
Quan Huynh-Thu et~al.
\newblock Scope of validity of psnr in image/video quality assessment.
\newblock {\em Electronics letters}, 44(13):800--801, 2008.

\bibitem{isola2017image}
Phillip Isola et~al.
\newblock Image-to-image translation with conditional adversarial networks.
\newblock In {\em Proceedings of the IEEE conference on computer vision and pattern recognition}, pages 1125--1134, 2017.

\bibitem{Jocher_Ultralytics_YOLO_2023}
Glenn Jocher et~al.
\newblock {Ultralytics YOLO}, Jan. 2023.

\bibitem{kansal2024privacy}
Kajal Kansal et~al.
\newblock Privacy-enhancing person re-identification framework-a dual-stage approach.
\newblock In {\em Proceedings of the IEEE/CVF Winter Conference on Applications of Computer Vision}, pages 8543--8552, 2024.

\bibitem{kappel2021high}
Moritz Kappel et~al.
\newblock High-fidelity neural human motion transfer from monocular video.
\newblock In {\em Proceedings of the IEEE/CVF conference on computer vision and pattern recognition}, pages 1541--1550, 2021.

\bibitem{li2021pose}
Ke Li et~al.
\newblock Pose recognition with cascade transformers.
\newblock In {\em Proceedings of the IEEE/CVF conference on computer vision and pattern recognition}, pages 1944--1953, 2021.

\bibitem{li2021tokenpose}
Yanjie Li et~al.
\newblock Tokenpose: Learning keypoint tokens for human pose estimation.
\newblock In {\em Proceedings of the IEEE/CVF International conference on computer vision}, pages 11313--11322, 2021.

\bibitem{lin2014microsoft}
Tsung-Yi Lin et~al.
\newblock Microsoft coco: Common objects in context.
\newblock In {\em Computer Vision--ECCV 2014: 13th European Conference, Zurich, Switzerland, September 6-12, 2014, Proceedings, Part V 13}, pages 740--755. Springer, 2014.

\bibitem{liu2020indoor}
Jixin Liu et~al.
\newblock Indoor privacy-preserving action recognition via partially coupled convolutional neural network.
\newblock In {\em 2020 International Conference on Artificial Intelligence and Computer Engineering (ICAICE)}, pages 292--295. IEEE, 2020.

\bibitem{liu2021motion}
Zhenguang Liu et~al.
\newblock Motion prediction using trajectory cues.
\newblock In {\em Proceedings of the IEEE/CVF international conference on computer vision}, pages 13299--13308, 2021.

\bibitem{lopez2024privacy}
Jhon Lopez et~al.
\newblock Privacy-preserving optics for enhancing protection in face de-identification.
\newblock In {\em Proceedings of the IEEE/CVF Conference on Computer Vision and Pattern Recognition}, pages 12120--12129, 2024.

\bibitem{luo2021rethinking}
Zhengxiong Luo et~al.
\newblock Rethinking the heatmap regression for bottom-up human pose estimation.
\newblock In {\em Proceedings of the IEEE/CVF conference on computer vision and pattern recognition}, pages 13264--13273, 2021.

\bibitem{mao2022poseur}
Weian Mao et~al.
\newblock Poseur: Direct human pose regression with transformers.
\newblock In {\em European conference on computer vision}, pages 72--88. Springer, 2022.

\bibitem{maximov2020ciagan}
Maxim Maximov et~al.
\newblock Ciagan: Conditional identity anonymization generative adversarial networks.
\newblock In {\em Proceedings of the IEEE/CVF conference on computer vision and pattern recognition}, pages 5447--5456, 2020.

\bibitem{mirza2014conditional}
Mehdi Mirza et~al.
\newblock Conditional generative adversarial nets.
\newblock {\em arXiv preprint arXiv:1411.1784}, 2014.

\bibitem{NvidiaJetsonOrin2023}
{NVIDIA Corporation}.
\newblock Nvidia jetson agx orin developer kit user guide.
\newblock Available from NVIDIA, 2023.

\bibitem{padilla2015visual}
Jos{\'e}~Ram{\'o}n Padilla-L{\'o}pez et~al.
\newblock Visual privacy protection methods: A survey.
\newblock {\em Expert Systems with Applications}, 42(9):4177--4195, 2015.

\bibitem{paolanti2018person}
Marina Paolanti et~al.
\newblock Person re-identification with rgb-d camera in top-view configuration through multiple nearest neighbor classifiers and neighborhood component features selection.
\newblock {\em Sensors}, 18(10):3471, 2018.

\bibitem{paszke2019pytorch}
Adam Paszke et~al.
\newblock Pytorch: An imperative style, high-performance deep learning library.
\newblock {\em Advances in neural information processing systems}, 32, 2019.

\bibitem{pishchulin2013poselet}
Leonid Pishchulin et~al.
\newblock Poselet conditioned pictorial structures.
\newblock In {\em Proceedings of the IEEE conference on computer vision and pattern recognition}, pages 588--595, 2013.

\bibitem{ren2018learning}
Zhongzheng Ren et~al.
\newblock Learning to anonymize faces for privacy preserving action detection.
\newblock In {\em Proceedings of the european conference on computer vision (ECCV)}, pages 620--636, 2018.

\bibitem{ronneberger2015u}
Olaf Ronneberger et~al.
\newblock U-net: Convolutional networks for biomedical image segmentation.
\newblock In {\em Medical image computing and computer-assisted intervention--MICCAI 2015: 18th international conference, Munich, Germany, October 5-9, 2015, proceedings, part III 18}, pages 234--241. Springer, 2015.

\bibitem{ruan2019devil}
Tao Ruan et~al.
\newblock Devil in the details: Towards accurate single and multiple human parsing.
\newblock In {\em Proceedings of the AAAI conference on artificial intelligence}, volume~33, pages 4814--4821, 2019.

\bibitem{ryoo2017privacy}
Michael Ryoo et~al.
\newblock Privacy-preserving human activity recognition from extreme low resolution.
\newblock In {\em Proceedings of the AAAI conference on artificial intelligence}, volume~31, 2017.

\bibitem{sarkar2022secure}
Nurul~I Sarkar et~al.
\newblock A secure long-range transceiver for monitoring and storing iot data in the cloud: design and performance study.
\newblock {\em Sensors}, 22(21):8380, 2022.

\bibitem{shi2022end}
Dahu Shi et~al.
\newblock End-to-end multi-person pose estimation with transformers.
\newblock In {\em Proceedings of the IEEE/CVF Conference on Computer Vision and Pattern Recognition}, pages 11069--11078, 2022.

\bibitem{srivastav2019human}
Vinkle Srivastav et~al.
\newblock Human pose estimation on privacy-preserving low-resolution depth images.
\newblock In {\em International conference on medical image computing and computer-assisted intervention}, pages 583--591. Springer, 2019.

\bibitem{teepe2022towards}
Torben Teepe et~al.
\newblock Towards a deeper understanding of skeleton-based gait recognition.
\newblock In {\em Proceedings of the IEEE/CVF conference on computer vision and pattern recognition}, pages 1569--1577, 2022.

\bibitem{tompson2015efficient}
Jonathan Tompson et~al.
\newblock Efficient object localization using convolutional networks.
\newblock In {\em Proceedings of the IEEE conference on computer vision and pattern recognition}, pages 648--656, 2015.

\bibitem{toshev2014deeppose}
Alexander Toshev et~al.
\newblock Deeppose: Human pose estimation via deep neural networks.
\newblock In {\em Proceedings of the IEEE conference on computer vision and pattern recognition}, pages 1653--1660, 2014.

\bibitem{wang2013beyond}
Fang Wang et~al.
\newblock Beyond physical connections: Tree models in human pose estimation.
\newblock In {\em Proceedings of the IEEE conference on computer vision and pattern recognition}, pages 596--603, 2013.

\bibitem{wang2013action}
Heng Wang et~al.
\newblock Action recognition with improved trajectories.
\newblock In {\em Proceedings of the IEEE international conference on computer vision}, pages 3551--3558, 2013.

\bibitem{wang2022regularizing}
Haixin Wang et~al.
\newblock Regularizing vector embedding in bottom-up human pose estimation.
\newblock In {\em European Conference on Computer Vision}, pages 107--122. Springer, 2022.

\bibitem{wang2020graph}
Jian Wang et~al.
\newblock Graph-pcnn: Two stage human pose estimation with graph pose refinement.
\newblock In {\em Computer Vision--ECCV 2020: 16th European Conference, Glasgow, UK, August 23--28, 2020, Proceedings, Part XI 16}, pages 492--508. Springer, 2020.

\bibitem{wang2022lite}
Yihan Wang et~al.
\newblock Lite pose: Efficient architecture design for 2d human pose estimation.
\newblock In {\em Proceedings of the IEEE/CVF Conference on Computer Vision and Pattern Recognition}, pages 13126--13136, 2022.

\bibitem{wang2004image}
Zhou Wang et~al.
\newblock Image quality assessment: from error visibility to structural similarity.
\newblock {\em IEEE transactions on image processing}, 13(4):600--612, 2004.

\bibitem{xiao2020adversarial}
Taihong Xiao et~al.
\newblock Adversarial learning of privacy-preserving and task-oriented representations.
\newblock In {\em Proceedings of the AAAI Conference on Artificial Intelligence}, volume~34, pages 12434--12441, 2020.

\bibitem{xu2021action}
Feiyi Xu et~al.
\newblock Action recognition framework in traffic scene for autonomous driving system.
\newblock {\em IEEE Transactions on Intelligent Transportation Systems}, 23(11):22301--22311, 2021.

\bibitem{yang2021transpose}
Sen Yang et~al.
\newblock Transpose: Keypoint localization via transformer.
\newblock In {\em Proceedings of the IEEE/CVF international conference on computer vision}, pages 11802--11812, 2021.

\bibitem{yang2011articulated}
Yi Yang et~al.
\newblock Articulated pose estimation with flexible mixtures-of-parts.
\newblock In {\em CVPR 2011}, pages 1385--1392. IEEE, 2011.

\bibitem{yao2010modeling}
Bangpeng Yao et~al.
\newblock Modeling mutual context of object and human pose in human-object interaction activities.
\newblock In {\em 2010 IEEE Computer Society Conference on Computer Vision and Pattern Recognition}, pages 17--24. IEEE, 2010.

\bibitem{yogameena2017computer}
Balasubramanian Yogameena et~al.
\newblock Computer vision based crowd disaster avoidance system: A survey.
\newblock {\em International journal of disaster risk reduction}, 22:95--129, 2017.

\bibitem{yu2024advanced}
Chang Yu et~al.
\newblock Advanced user credit risk prediction model using lightgbm, xgboost and tabnet with smoteenn.
\newblock {\em arXiv preprint arXiv:2408.03497}, 2024.

\bibitem{yu2024credit}
Chang Yu et~al.
\newblock Credit card fraud detection using advanced transformer model.
\newblock {\em arXiv preprint arXiv:2406.03733}, 2024.

\bibitem{yun2024hypersense}
Sanggeon Yun et~al.
\newblock Hypersense: Hyperdimensional intelligent sensing for energy-efficient sparse data processing.
\newblock {\em Advanced Intelligent Systems}, page 2400228, 2024.

\bibitem{yun2024missiongnn}
Sanggeon Yun et~al.
\newblock Missiongnn: Hierarchical multimodal gnn-based weakly supervised video anomaly recognition with mission-specific knowledge graph generation.
\newblock {\em arXiv preprint arXiv:2406.18815}, 2024.

\bibitem{zhao2022freed}
Bowen Zhao et~al.
\newblock Freed: An efficient privacy-preserving solution for person re-identification.
\newblock In {\em 2022 IEEE Conference on Dependable and Secure Computing (DSC)}, pages 1--8. IEEE, 2022.

\bibitem{zheng2024advanced}
Qi Zheng et~al.
\newblock Advanced payment security system:xgboost, lightgbm and smote integrated.
\newblock {\em arXiv preprint arXiv:2406.04658}, 2024.

\end{thebibliography}
}

\end{document}